# A DEEP NEURAL NETWORKS APPROACH FOR PIXEL-LEVEL RUNWAY PAVEMENT CRACK SEGMENTATION USING DRONE-CAPTURED IMAGES

**Liming Jiang**, Ph.D. Student
Department of Civil and Environmental Engineering
University of Massachusetts Lowell
One University Ave, Lowell, MA 01854
Email: Liming_Jiang@student.uml.edu

**Yuanchang Xie***, Ph.D., P.E.
Associate Professor
Department of Civil and Environmental Engineering
University of Massachusetts Lowell
One University Ave, Lowell, MA 01854
Phone: (978) 934-3681
Email: Yuanchang_Xie@uml.edu

**Tianzhu Ren**, Ph.D.
Software Engineer
Amazon
440 Terry Ave N, Seattle, WA 98109
Email: rtianzhu@amazon.com

* Corresponding Author

## ABSTRACT

Pavement conditions are a critical aspect of asset management and directly affect safety.  This study introduces a deep neural network method called U-Net for pavement crack segmentation based on drone-captured images to reduce the cost and time needed for airport runway inspection.  The proposed approach can also be used for highway pavement conditions assessment during off-peak periods when there are few vehicles on the road.  In this study, runway pavement images are collected using drone at various heights from the Fitchburg Municipal Airport (FMA) in Massachusetts to evaluate their quality and applicability for crack segmentation, from which an optimal height is determined.  Drone images captured at the optimal height are then used to evaluate the crack segmentation performance of the U-Net model.  Deep learning methods typically require a huge set of annotated training datasets for model development, which can be a major obstacle for their applications.  An online annotated pavement image dataset is used together with the FMA data to train the U-Net model.  The results show that U-Net performs well on the FMA testing data even with limited FMA training images, suggesting that it has good generalization ability and great potential to be used for both airport runways and highway pavements.

## INTRODUCTION

Drones are being explored by many state Departments of Transportation (DOTs) for innovative applications, including highway asset inspections.  This research focuses on investigating the potential of using drone mounted cameras for pavement conditions assessment, which was traditionally conducted by manual methods and recently mostly by vehicle mounted camera and laser sensors.  Vehicle based inspection platforms make it possible to use highly precise but bulky sensors like laser. However, they usually can only cover one lane in a single run.  Drones are emerging as a cost-effective solution for many asset inspection applications.  Some of them can travel at very high speeds and cover large areas in a short amount of time.  Beyond a certain height, they can cover a wide path (e.g., covering multiple lanes).  However, drones also have constraints such as short flight time and limited payload, especially for those relatively low-cost drones.  Their capability for pavement conditions assessment has not been fully investigated and understood, particularly in terms of the quality of drone images and the feasibility of extracting useful pavement distress information out of them.

Recent advances in computer vision and deep learning have led to the development of many deep learning applications in image processing. In light of the promising performance of deep learning models, this pilot study applies a U-Net (*1,2*) deep neural networks model to analyze airport runway pavement images captured by drones, and to explore its applicability in pavement inspection even with a limited training dataset.  The rest of the paper is organized as follows: the next section provides a review of some most recent studies on deep learning and pavement crack detection.  Following that is a brief description of the adopted U-Net model, the data used in this study and how the U-Net model is applied, after which the U-Net modeling results are presented.  Finally, conclusions and discussion are provided.

## LITERATURE REVIEW

Deep learning has significantly improved the analysis accuracy of image processing and enabled numerous innovative transportation applications. This review focuses on some most relevant studies on pavement crack detection and provides an overview of the state-of-the-art works. Zhang et al. (*3*) developed a CNN model to detect pavement cracks and tested it based on a dataset captured by smartphone. The input images were divided into 90 × 90 pixel patches and classified as either cracked or non-cracked patches. The CNN generated more accurate results than support vector machines (SVM) and a Boosting method. Based on the same images used in Zhang et al. (*3*), Pauly et al. (*4*) considered a deeper CNN structure. Their CNN consisted of four convolutional layers, four subsampling or max pooling layers, and two fully connected layers. Pauly et al. concluded that the deeper CNN architecture did increase pavement distress detection success, but did not present strong results when tested on images from new locations. Eisenbech et al. (*5*) established the open source GAPS (German Asphalt Pavement Distress) database, which consisted of 1,969 grayscale images. Similar to Pauly et al. (*4*), Eisenbech et al. later considered a deeper network structure which consisted of eight convolutional layers, three subsampling or max pooling layers, and three fully connected layers. In general, the deeper CNN achieved higher generalizability when employing the dropout or batch normalization techniques to avoid overfitting. Maeda et al. (*6*) developed a smartphone App called RoadDamageDetector to detect pavement distress. They utilized a smartphone mounted on car dashboard to collect data and distinguished between eight types of pavement distress. The Single Shot MultiBox Detector (SSD) with Inception V2 and SSD using MobileNet frameworks were utilized in their study. The RoadDamageDetector App achieved a precision of greater than 75%. Fan et al. (*7*) used images from two publicly available databases, the CFD database and the AigleRN database. They developed a CNN with four convolutional layers, two subsampling or max pooling layers, and three fully connected layers. They first trained the network on the CFD database and tested it on the AigleRN database. Then, they trained the CNN on the AigleRN database and tested it on the CFD database. Finally, they trained and tested the CNN on images from both the CFD and AigleRN databases. The results showed good generalizability, especially when the network was trained on the hybrid database. Rather than developing a CNN from scratch, Gopalakrishnan et al. (*8*) applied transfer learning to a pre-trained CNN. They utilized the VGG-16 CNN, which was pre-trained on the ImageNet database. They truncated the VGG-16 CNN, and only used the convolutional layers of the pre-trained CNN but not the fully connected layers. Then, a new fully connected layer was trained on the features recorded by the truncated VGG-16 CNN. They utilized 1,056 images from the FHWA's publicly available pavement performance database. The best results were generated by a single layer classifier which was trained on the pre-trained VGG-16 CNN. Gopalakrishnan et al. also showed the potential of utilizing pre-trained VGG-16 CNN model with transfer learning for pavement distress detection based on drone collected images and achieved an accuracy of 89% in crack detection (*9,10*). Cha et al. (*11*) reported using MatConvNet, an open source Matlab toolbox for developing CNN, for the task of crack detection. The accuracies in the training and validation phases were 98.22% and 97.95%, respectively. Cha et al. also found that their CNN significantly outperformed the Canny and Sobel edge detectors for identifying pavement cracks.

The above studies all considered pavement crack detection at an aggregate level, instead of the pixel level. They divided the input images into small patches (e.g., 90 by 90 pixels) and classify each patch as either cracked or non-cracked. Such results cannot provide the information needed by detailed pavement conditions analyses such as generating Pavement Condition Index (PCI). Some more recent studies explored how to classify each pixel as cracked or non-cracked. Zhang et al. (*12*) proposed a CrackNet based on CNN that can detect cracks at the pixel level using 3D asphalt surface data. CrackNet was shown to significantly outperform SVM. Zhang et al. (*13*) further improved the CrackNet by adopting a recurrent neural network (RNN) structure and named it as CrackNet-R. Yang et al. (*14*) developed a fully convolutional network (FCN) and compared it with CrackNet. They concluded that CrackNet was better at detecting thin cracks, although FCN needed less training time. Ji et al. (*15*) applied U-Net to detect cracks in concrete structures and found it to outperform Canny and Sobel edge detection methods. Cheng et al. (*16*) applied U-Net to the CFD and AigleRN pavement crack databases separately, and found it to perform much better than CrackTree, CrackIT and CrackForest. However, both Jie et al. (*15*) and Cheng et al. (*16*) did not evaluate the generalization ability of U-Net. Recently, Yang et al. (*17*) proposed a feature pyramid and hierarchical boosting network (FPHBN) for pavement crack detection. They tested it using a CRACK500 pavement image database and concluded that FPHBN outperforms CrackForest and several CNN methods.

In summary, deep neural networks have attracted tremendous attention in the past few years for image-based crack detection. Earlier studies mostly divided input images into small patches and classified each patch as either cracked or non-cracked. Several recent studies focused on detecting cracks at the pixel level. Among these studies, Zhang et al. (*12,13*) used 3D asphalt surface data, which is different from the 2D data used in this research. In addition, it is difficult to capture accurate 3D surface data using drone due to sensor weight and payload constraint. Although some researchers (*15,16*) have tried U-Net, none of them applied it to drone collected pavement images and both did not consider U-Net's generalization ability. Also, one of the studies (*15*) was about concrete structures, not pavement cracks.

## METHODOLOGY

### U-Net

This research utilizes a promising deep learning approach called U-Net (*1,2*) for drone collected pavement image analysis. U-Net is a pixel-level classifier and was initially proposed for biomedical image segmentation. Its name is from the "U" shape architecture shown in Figure 1. It utilizes a contracting path to capture context, which is followed by a symmetric expanding path for localization. The main advantage of U-Net is that it does not require tedious image preprocessing and feature engineering work and can be applied in an end-to-end fashion. Also, by using data augmentation techniques U-Net can achieve decent prediction performance with a small set of annotated training images. It has been proven to perform faster and better than many prior best models at the IEEE International Symposium on Biomedical Imaging (ISBI) Challenge for segmentation of neuronal structures.

Some hyperparameter tunings on the original U-Net has been performed before applying it to analyzing drone collected runway pavement images. Inspired by (*4,5*), a deeper structure is considered and the number of channels at each convolutional layer is increased 0.5 times to improve model fitting and generalization ability. Additionally, the image input dimension is set to $256 \times 256$ pixels.

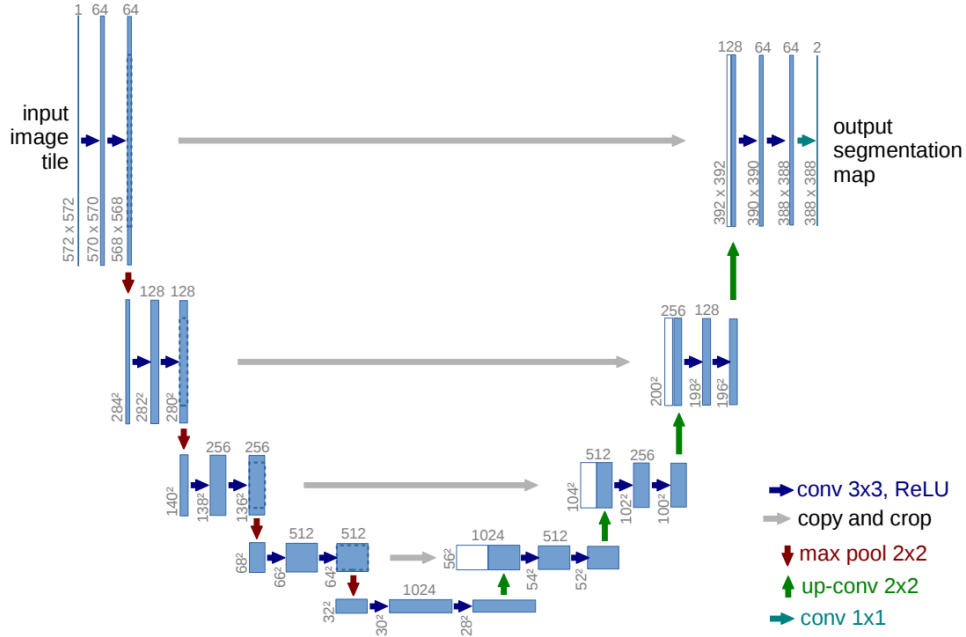

**Figure 1: U-Net Architecture (*1,2*)**

### Measurement of Performance

Crack segmentation is essentially a classification task at the pixel level. Furthermore, it is a difficult classification problem due to the highly imbalanced instances, i.e., much less crack pixels than non-crack pixels. To accurately measure the performance of the proposed model, the following *F1 score*, *Precision* and *Recall* metrics are presented first, and later an *Intersection over Union* (IoU) is introduced and finally used in this research.

$$F1 = 2 * \frac{Precision * Recall}{Precision + Recall} \quad (1)$$

$$Precision = \frac{TP}{TP + FP} \quad (2)$$

$$Recall = \frac{TP}{TP + FN} \quad (3)$$

where *TP* denotes *True Positive*, *FP* denotes *False Positive*, and *FN* denotes *False Negative*. *F1* considers both *Precision* and *Recall*, and is calculated as the harmonic average of *Precision* and *Recall*. *F1* score value ranges between 0 and 1. Larger *F1* scores mean better model performances.

Many previous classification studies consider *Precision* as the main performance metric, which can be problematic. *Precision* may work if different outcome categories have approximately the same number of observations. However, pavement crack datasets are often characterized by a highly imbalanced class distribution. Based solely on the *Precision* performance metric, the best-performing model may favor non-crack pixels and generate poor results for crack pixels. Such a model does not provide much useful information for crack detection and may lead to biased if not erroneous conclusions. Therefore, *F1* (see Eq. (1)) is often used in recent studies.

In this research, another intuitive and informative metric, *IoU* (*18*), is introduced and adopted. As shown in Figure 2, *IoU* is defined as the area of intersection (green area) divided by the area of union (red area). Intersection represents the area covered by both the ground truth and prediction, while union includes the area in either the ground truth or prediction. If the prediction matches the ground truth perfectly, then the corresponding *IoU* would be 100%.

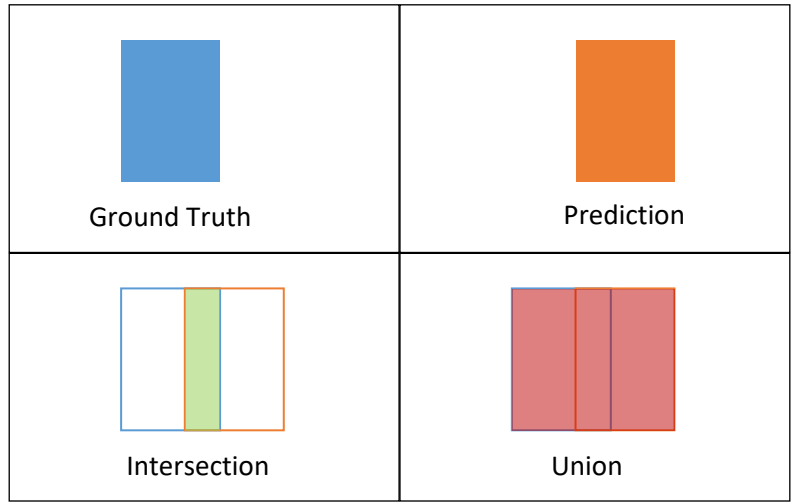

**Figure 2: Definition of IoU**

**Data**

**Table 1: Flight Plan for the Second Drone Data Collection**

| Flight Description | System | Height (ft) | Coverage |
|---|---|---|---|
| Elevator Flight | DJI Matrice 210/XT2 Dual | 10-70 (at a 5 ft interval), 80-150 (a 10 ft interval), and 200 | Single Point |
| Coverage Flight | DJI Inspire 2/X4S | 32, 40, 50, and 60 | Half Runway |
| | DJI Inspire 2/X4S | 70, 90, 120, and 200 | Full Runway |

This research first investigated how drone height may affect the image quality. Table 1 shows how the investigation/data collection was conducted. The *elevator* flight captured pavement images at a single location but different heights. The purpose was to find out how height may affect the image quality. With such information, the team was hoping to identify a threshold height, beyond which the collected data may become much less useful. Note that such a threshold may depend on the types of drone and camera used. Therefore, the conclusion in this study may not be generalized to other cameras and drones. The *coverage* flight was designed to collect additional runway pavement images for further crack detection model development and testing.

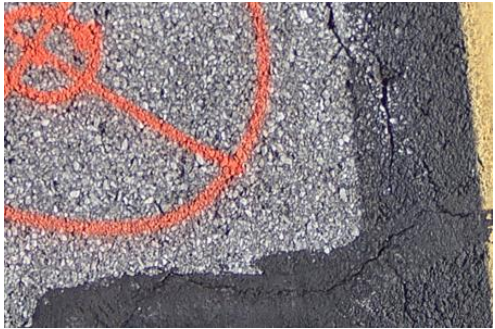
10 ft

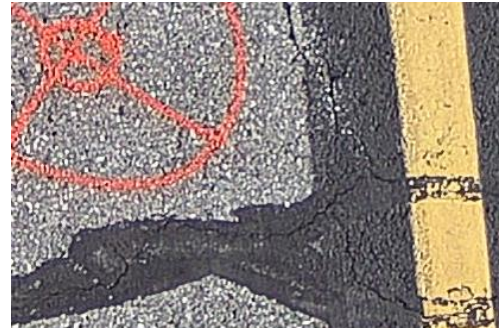
40 ft

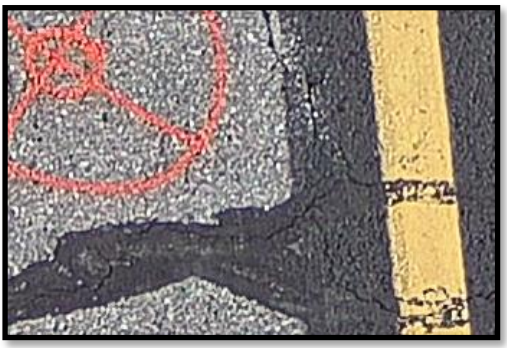
50 ft

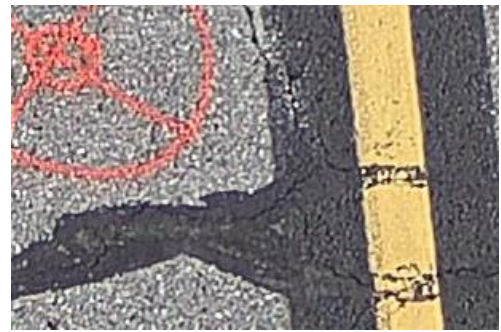
60 ft

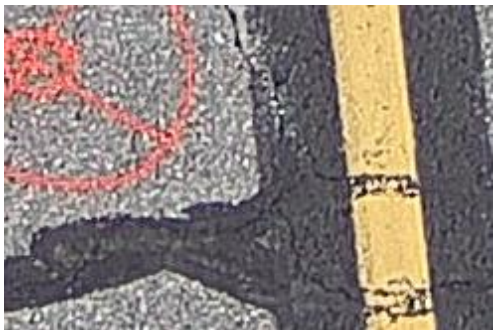
70 ft

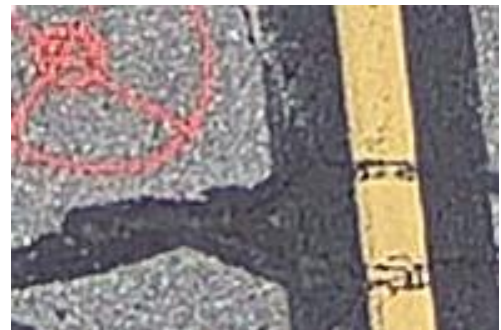
120 ft

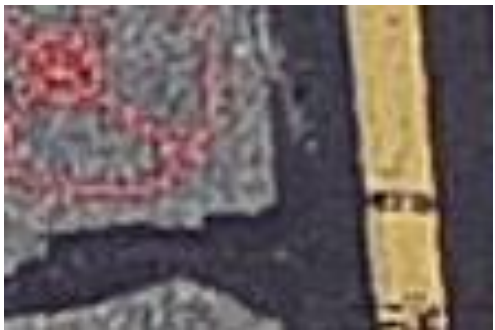
150 ft

**Figure 3: Pavement Image Taken by the Elevator Flight**

Figure 3 shows samples of pavement images collected at different heights through the elevator flight. It can be seen that once the height is greater than 120 ft, the pavement crack details are very difficult to see. Choosing the drone flight height needs to consider both the image resolution and the efficiency. For instance, the FMA runway is about 150 ft wide. At the 10 ft height, it will take a drone 10 runs to cover the entire width of the runway (each run covers a width of 15 ft). While at the 50 ft height, the drone just need 2 runs (a width of 78 ft for each run). Again, depending on the cameras used, this conclusion may be different. However, the overall trend will be the same: higher heights will require less flights (i.e., better efficiency) but result in poorer image resolutions. To achieve a balance between image quality and efficiency, the research decided to use the images captured at 50 ft.

Table 2: Data Used in U-Net Modeling

| Dataset Name | Total Images | Training Images | Testing Images |
|---|---|---|---|
| FMA | 160 | 70 | 90 |
| Crack500 | 2,244 | 1,896 | 348 |

This research also obtained an additional Crack500 dataset (*3,17*) collected by a group at the Temple University using smartphone. The Crack500 dataset contains both raw pavement images and annotated images. The annotated images mark cracks at the pixel level and are essential for training/teaching neural networks models how to detect cracks. Preparing the annotated images is very time-consuming. Therefore, it is beneficial to have datasets that include both the original and annotated images. The drone images (5472 pixels x 3648 pixels) obtained at the 50 ft height from the FMA were further divided into smaller pieces (256 pixels x 256 pixels). These new images were then annotated manually (i.e., mark those crack pixels one by one) and used for model training and testing. Table 2 shows how the input Crack500 and FMA data was separated into training and testing datasets.

**Model Application**

Data augmentation was employed to further increase the amount of training data for the U-Net model. For hyperparameters tuning, The Adam optimizer was utilized with a learning rate of .0001. The number of training episodes was set to 1000, and binary cross entropy was adopted as the loss function. ReLu activation function was used for all layers except for the last one, which utilized the sigmoid function. Additionally, the batch size was set to 5.

Since deep learning methods typically require a substantial number of annotated training images, it is very significant to investigate whether data from one source can be used to train a model that will be applied to data from a different source. For this purpose, the Crack500 and FMA datasets were both used for the U-Net model training. Two U-Net models were developed to find out whether a U-Net model trained on the Crack500 dataset can perform well on the FMA dataset. The first U-Net$_{Crack500}$ model was trained solely based on the Crack500 dataset using 1,896 images (See Table 2). The second U-Net$_{Crack500\&FMA}$ model was trained using 1,896 Crack500 and 70 FMA images.

The U-Net$_{Crack500}$ model was initially evaluated based on the Crack500 data using the remaining 348 images set aside for testing, and its performance was very promising. This model was further tested on the FMA data and resulted in less accurate performance, which is not surprising and could be attributed to the differences between the two datasets. The Crack500 dataset was collected using a camera about 7 ft above the ground, while the FMA dataset was collected 50 ft above the ground. Given the limited size of the FMA dataset, training a U-Net model solely based on it is difficult. Therefore, the U-Net$_{Crack500\&FMA}$ model was developed and tested on the FMA data. The detailed model evaluation results are presented in the following section.

**RESULTS ANALYSIS**

Table 3 presents the performance results of the two U-Net models and compares them with the results reported in a 2019 research paper by Yang et al. (*17*) that also used the Crack500 data. From the evaluation results based on the Crack500 data, U-Net clearly outperforms (*IoU*=0.60) the benchmark FPHBN model (*17*).

However, the Crack500 data was collected using smartphone, while the FMA data was captured using a drone 50 ft above the runway. Due to the differences between the Crack500 and FMA datasets, applying the U-Net$_{Crack500}$ model directly to the FMA data did not generate satisfactory results (*IoU*=0.30). By combining only 70 annotated images from the FMA dataset with the Crack500 data for model training, the U-Net$_{Crack500\&FMA}$ model was able to produce significantly better performance (*IoU*=0.56) compared to the U-Net$_{Crack500}$ model on the FMA testing data.

**Table 3: Model Performance Comparison**

| Evaluated based on | Model | Performance (*IoU*) |
|---|---|---|
| Crack500 | Benchmark FPHBN model (*17*) | 0.49 |
| | U-Net$_{Crack500}$ | 0.60 |
| FMA | U-Net$_{Crack500}$ | 0.30 |
| | U-Net$_{Crack500\&FMA}$ | 0.56 |

The *IoU* results are useful for model comparision, since large values represent better performances. In addition, Figure 4 is included to visually illustrate the prediction performances of the U-Net$_{Crack500}$ model on some randomly selected Crack500 testing data. Images from the first row are the original pavement data; images in the second row are the ground truth (i.e., true cracks); and images in the last row are the prediction results generated by the U-Net$_{Crack500}$ model. Comparing the ground truth and prediction results in Figure 4 suggests that the U-Net$_{Crack500}$ model performs very well and is able to accurately capture all major cracks.

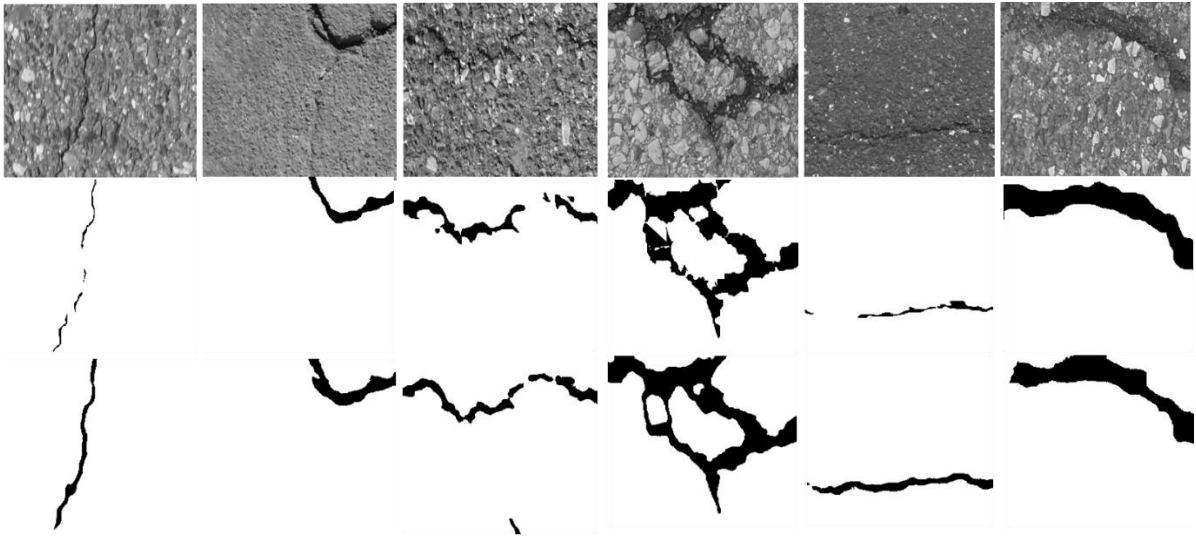

**Figure 4: Prediction Results of U-Net$_{Crack500}$ on Crack500 Data (1$^{st}$ Row – Original Images, 2$^{nd}$ Row – Ground Truth, and 3$^{rd}$ Row - Predictions)**

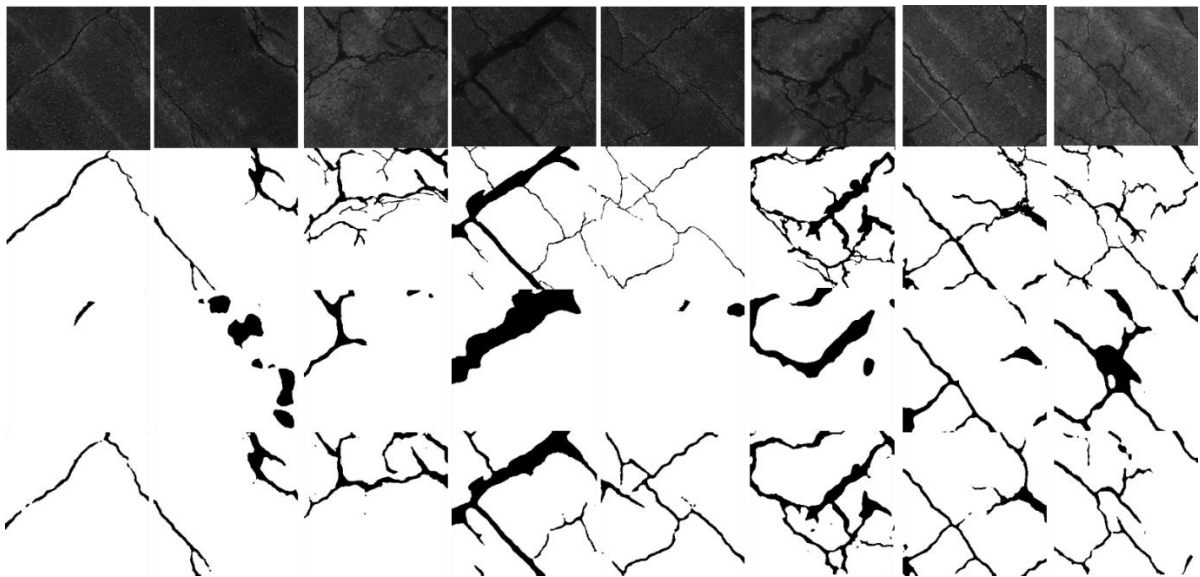

**Figure 5: Prediction Results of U-Net$_{Crack500}$ and U-Net$_{Crack500\&FMA}$ on FMA Data (1$^{st}$ Row – Original Images, 2$^{nd}$ Row – Ground Truth, 3$^{rd}$ Row - U-Net$_{Crack500}$ Predictions, 4$^{th}$ Row - U-Net$_{Crack500\&FMA}$ Predictions)**

Similarly, the prediction results of the U-Net$_{Crack500}$ and U-Net$_{Crack500\&FMA}$ models on the FMA testing data are further illustrated in Figure 5. A comparison of Figures 4 and 5 clearly suggests that the Crack500 and FMA datasets are very different. Crack500 data has a higher resolution and differences between cracks and non-cracks can be easily distinguished. On the other hand, FMA data overall is darker and has a lower resolution. Some of the cracks are difficult to identify even manually with human intelligence. In Figure 5, the third and fourth rows show the results of the U-Net$_{Crack500}$ and U-Net$_{Crack500\&FMA}$ models, respectively. Compared to the ground truth (i.e., the 2nd row), the U-Net$_{Crack500\&FMA}$ model clearly is able to identify more cracks than the U-Net$_{Crack500}$ model. Given the limited time, this research only annotated 160 FMA images. If more annotated FMA images are used in U-Net training, its prediction performance is expected to be further improved.

Additionally, the developed U-Net$_{Crack500}$ model was applied to some pavement images generated by a laser pavement scanning system. Such a system is usually mounted on a vehicle (about 2 meters/6.6 ft above the ground) and similar systems are widely used by state departments of transportation to collect highway pavement condition data. The testing results are presented in Figure 6. The highway pavement images have not been annotated and the comparison is directly between the raw images and the predicted results. The highway pavement images are taken about 2 meters/6.6 ft above the ground and are closer to the Crack500 data in terms of resolution than the FMA data. This probably explains why the U-Net$_{Crack500}$ model's performance on this dataset is very encouraging (compared to the direct application results on the FMA data), although it is trained on a different set of images. Overall, this suggests that the U-Net$_{Crack500}$ model has great generalization ability and can potentially be used directly by highway departments on datasets that are similar to Crack500.

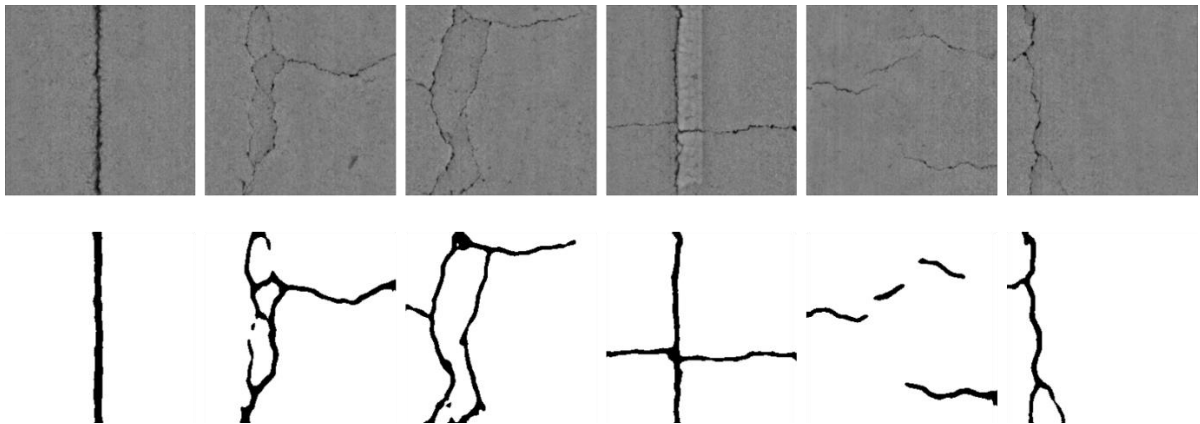

**Figure 6: Prediction Results of U-Net$_{Crack500}$ on Highway Pavement Image Data (1st Row - Original Images, and 2nd Row - Predictions)**

## CONCLUSIONS AND DISCUSSION

This pilot study applied U-Net model to analyze airport runway pavement images collected by drone, and to explore the possibility of using a hybrid training dataset for model training to address the typical needs of deep learning methods for a large annotated training dataset and tremendous data preparation effort. Runway pavement images were collected from the Fitchburg Municipal Airport (FMA) and used together with the CRACK500 online pavement image dataset to train and test the U-Net model. The developed U-Net$_{Crack500}$ model was found to outperform a state-of-the-art FPHBN model (*17*) published in 2019, which was also trained and evaluated on the same CRACK500 dataset. A hybrid dataset consisting of images from CRACK500 and FMA was used to train another U-Net$_{Crack500\&FMA}$ model. With only 70 images from the FMA, this U-Net$_{Crack500\&FMA}$ model performed very well on the FMA testing images. This model was also applied to some images collected using a laser pavement scanning system, suggesting that the develop model has strong model generalization ability and great potential to be used for both airport runways and highway pavements.

Aside from its promising crack segmentation performance and strong generalization ability, the U-Net model provides an end-to-end solution to detect cracks at the pixel level, which is a critical and desirable feature for in-depth pavement conditions assessment such as calculating PCI values.

## ACKNOWLEDGMENT

The authors would like to thank Massachusetts Department of Transportation Aeronautics Division for their financial support and for providing important guidance during the course of this research. The contents of this paper only reflect the views of the authors, who are responsible for the facts and the accuracy of the data presented herein.

## AUTHOR CONTRIBUTION STATEMENT

The authors confirm contribution to the paper as follows: study conception and design: Yuanchang Xie; data collection: Yuanchang Xie, Liming Jiang; analysis and interpretation of results: Yuanchang Xie, Liming Jiang, Tianzhu Ren; draft manuscript preparation: Yuanchang Xie, Liming Jiang, Tianzhu Ren. All authors reviewed the results and approved the final version of the manuscript.